\documentclass[10pt,twocolumn,letterpaper]{article}

\usepackage[pagenumbers]{cvpr} 

\definecolor{darkred}{rgb}{0.7, 0.0, 0.0}

\newcommand{\model}{UltraFusion\xspace}
\newcommand{\dataset}{UltraFusion dataset\xspace}

\definecolor{cvprblue}{rgb}{0.21,0.49,0.74}
\usepackage[pagebackref,breaklinks,colorlinks,allcolors=cvprblue]{hyperref}

\usepackage[accsupp]{axessibility}
\usepackage{adjustbox}
\usepackage{multirow}
\usepackage{url}

\def\paperID{12582} 
\def\confName{CVPR}
\def\confYear{2025}

\title{UltraFusion: Ultra High Dynamic Imaging using Exposure Fusion}

\author{Zixuan Chen\textsuperscript{1, 3}\footnotemark[1] \quad Yujin Wang\textsuperscript{1}\footnotemark[1] \quad Xin 
 Cai\textsuperscript{2} \quad Zhiyuan You\textsuperscript{2} \quad Zheming Lu\textsuperscript{3} \quad Fan Zhang\textsuperscript{1}\\
 Shi Guo\textsuperscript{1} \quad  Tianfan Xue\textsuperscript{2,1}\\
\textsuperscript{1}Shanghai AI Laboratory \quad \textsuperscript{2}The Chinese University of Hong Kong \quad 
\textsuperscript{3}Zhejiang University
\\
\tt\small \{zxchen, zheminglu\}@zju.edu.cn, \{wangyujin, zhangfan, guoshi\}@pjlab.org.cn  \\
\tt\small caixin@link.cuhk.edu.hk, zhiyuanyou@foxmail.com, tfxue@ie.cuhk.edu.hk\\
}
\begin{document}

\twocolumn[{%
\renewcommand\twocolumn[1][]{#1}%
\maketitle
\vspace{-30pt}
\begin{center}
    \centering
    \captionsetup{type=figure}
    \includegraphics[width=\linewidth]{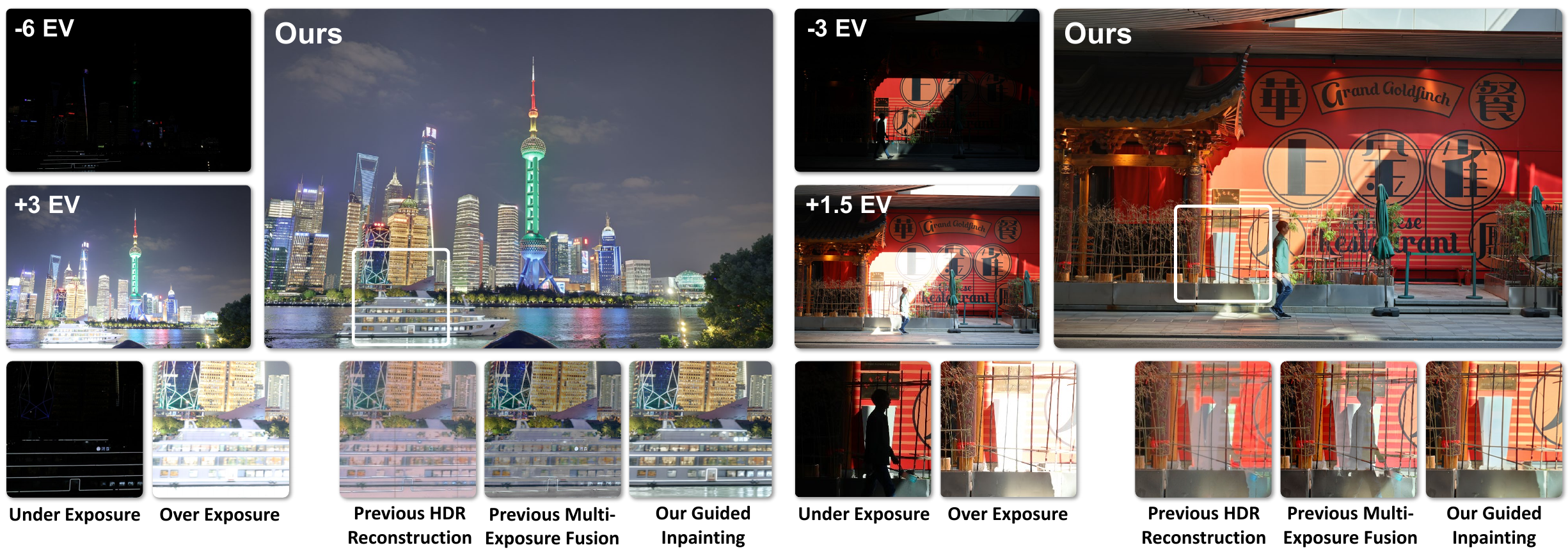}
    \captionof{figure}{Comparing our guided inpainting HDR imaging method with a state-of-the-art HDR reconstruction~\cite{liu2022hdrtransformer} and multi-exposure fusion~\cite{wu2024hsdsmef} methods. Both scenes are selected from our captured new real-world benchmark. \textbf{Left}: night cityscape with large exposure difference. \textbf{Right}: afternoon street with motion-induced occlusion. Previous methods struggle to handle these scenes. 
    By modeling HDR as an inpainting problem, our method can produce visually appealing results without ghosting artifacts in these challenging scenes.
    }
    \label{teaser}
\end{center}%
}]

\renewcommand{\thefootnote}{\fnsymbol{footnote}}
\footnotetext[1]{Equal contribution. This work was done during Zixuan Chen’s internship at Shanghai Artificial Intelligence Laboratory.}
\vspace{10pt}
\begin{abstract}


\vspace{-5pt}

Capturing high dynamic range (HDR) scenes is one of the most important issues in camera design. Majority of cameras use exposure fusion, which fuses images captured by different exposure levels, to increase dynamic range. However, this approach can only handle images with limited exposure difference, normally 3-4 stops. When applying to very high dynamic range scenes where a large exposure difference is required, this approach often fails due to incorrect alignment or inconsistent lighting between inputs, or tone mapping artifacts. In this work, we propose \model, the first exposure fusion technique that can merge inputs with 9 stops differences. The key idea is that we model exposure fusion as a \textbf{guided inpainting} problem, where the under-exposed image is used as a guidance to fill the missing information of over-exposed highlights in the over-exposed region. Using an under-exposed image as a soft guidance, instead of a hard constraint, our model is robust to potential alignment issue or lighting variations. Moreover, by utilizing the image prior of the generative model, our model also generates natural tone mapping, even for very high-dynamic range scenes. Our approach outperforms HDR-Transformer on latest HDR benchmarks. Moreover, to test its performance in ultra high dynamic range scenes, we capture a new real-world exposure fusion benchmark, \textit{\dataset}, with exposure differences up to 9 stops, and experiments show that \model~can generate beautiful and high-quality fusion results under various scenarios. Code and data will be available at \url{https://openimaginglab.github.io/UltraFusion}.

\end{abstract}
   
\vspace{-10pt}

\section{Introduction}
\label{sec:intro}

High dynamic range (HDR) imaging is one of the fundamental problems in the modern camera design. 
Due to hardware limitation, camera sensors have a much smaller dynamic range compared with the real-world. To increase it, majority of HDR solutions merge multiple images, either with the same~\cite{hasinoff2016burst} or different exposure levels~\cite{kalantari2017deep, wu2018deep, yan2019attention, yan2020deep, prabhakar2020towards, chen2022attention, liu2022ghost, song2022selective,tel2023alignment, kong2024safnet}. 
Despite all recent advances, majority HDR algorithms can only bring limited increase in dynamic range when input exposures are constrained.
For example, HDR+~\cite{hasinoff2016burst}, the first HDR algorithm used by commercial cameras, can only robustly increase the dynamic range by 8 times (3 stops). 
Therefore, in this work, we study the following question: can we drastically increase the dynamic range of a camera by fusing two images with very large exposure difference, like 9 stops in \cref{teaser}?

This is a fundamental challenging problem, due to the following three issues. First, to handle dynamic scenes, most of HDR fusion algorithms will first align input frames, which is very challenging when input has large brightness difference. As result, a ghosting issue happens when alignment fails, indicated by zoom-in patches in the right scene of~\cref{teaser}. Second, most of HDR algorithms assume that under-exposed image is simply a darker version of the normal image. However, the appearance of an object may change when exposure levels change, like ship in the left scene of~\cref{teaser}, resulting unnatural fusion result. Third, sometimes the result of fusion is an HDR image, which cannot be directly shown on normal low-dynamic-range display. Therefore, these HDR images will be further compressed through a tone-mapping process. When dynamic range is high, tone-mapping may introduce additional issue. Maintaining a natural contrast and rich details in the final output is challenging, as shown in the zoom-in patches of previous HDR reconstruction method in~\cref{teaser}.

\begin{figure}[t]
    \centering
    \includegraphics[width=\linewidth]{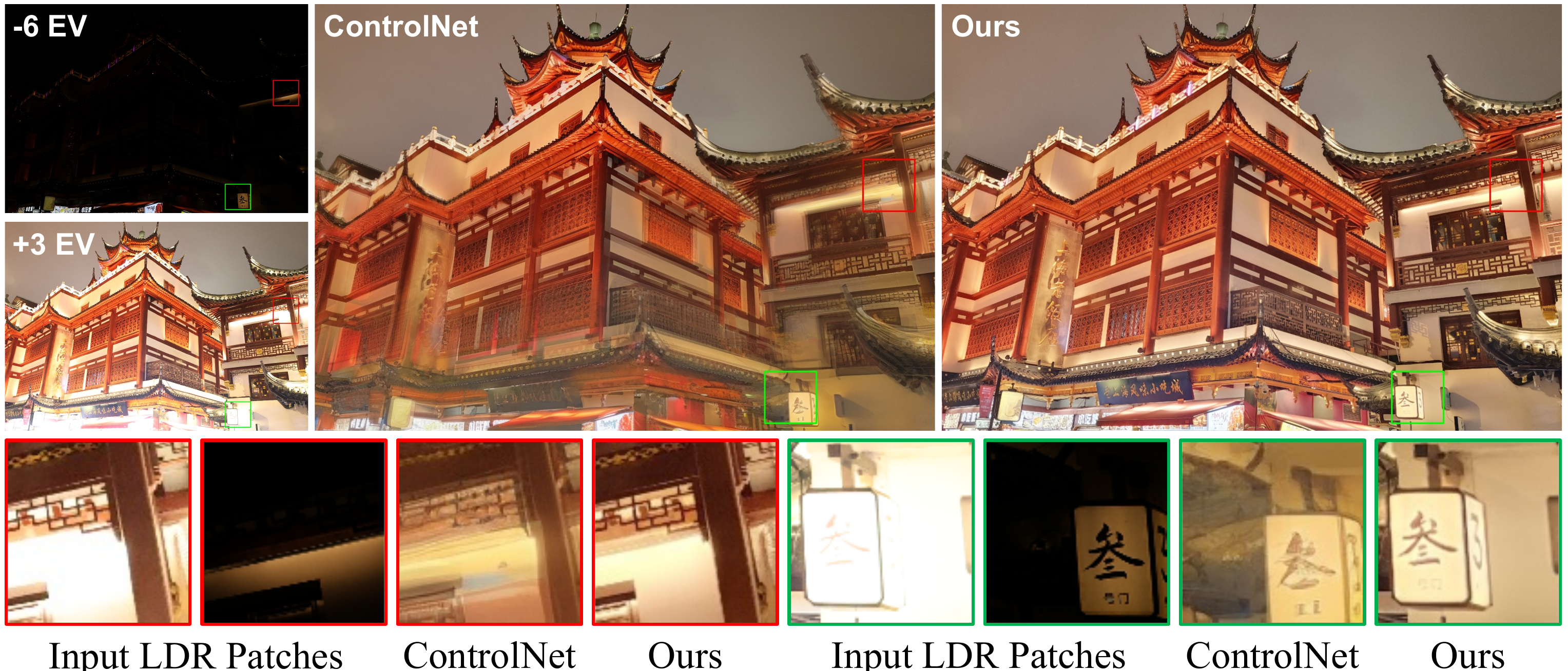}
    \caption{Visual comparison on directly utilizing ControlNet~\cite{controlnet} and our~\model. 
    Without pre-aligned data, ControlNet struggles to fix a frame for reference. However, our method fixes the over-exposed image as the reference and the under-exposed image as the guidance for inpainting, thereby avoiding artifacts.}
    \vspace{-15pt}
	\label{fig:Fig2}
\end{figure}


In this work, we propose a completely different fusion method, \model, which models it as a \emph{guided inpainting} problem. In this setup, the user captures two images, one normal exposed image where brighter objects are over-exposed, and another under exposed image, which only captures the very brighter parts of the scene. We use the normal exposed image as a reference, and inpainting the missing information in the highlight. Unlike the traditional inpainting, we use the information from under-exposed frame as guidance, so inpainted highlight is not completely generated, but stays consistent with under-exposed frame.

There are three advantages of this approach when handling large exposure difference. First, it follows the exposure fusion~\cite{mertens2007exposure} setup. Unlike HDR fusion techniques, which first generate HDR image and then compress to a low-dynamic-range (LDR) image through tone mapping, exposure fusion directly generates LDR output, avoiding cascading errors. That being said, our method outputs a tone-mapped LDR image instead of a linear HDR. Second, the under-exposed image is used as a soft guidance, instead of a hard constraint. Therefore, \model~is robust to alignment error (see the right scene of~\cref{teaser}) and lighting variation (see the left scene of ~\cref{teaser}). Third, image prior of the generative model ensures the natural look of the output image, reducing the potential artifacts. 

To train a guided inpainting, a simple solution is to train a ControlNet~\cite{controlnet} using two input images. However, such paradigm cannot handle dynamic HDR scenes, as ControlNet may not know which frame to choose as the reference, increasing the difficulty to fuse the information from over and short-exposed images. For instance, as shown in~\cref{fig:Fig2}, ControlNet selects the over-exposed image as the reference frame in red boxed region, but chooses the under-exposed image as the reference frame in green boxed region, leading to substantial artifacts in the result. Additionally, as a generative model, it may inevitably generate fake image content, as indicated by green box in~\cref{fig:Fig2}. 

To address these challenges, we design \model as follows. First, we warp the short-exposed image to the long-exposed one, and mask out the occluded regions. Then, we utilize the diffusion prior to inpaint the long-exposed image guided by partial short-exposed information. To make the network retain more details to generate guidance information, we propose a new decompose-and-fuse control branch, which eliminates the luminance component of the short-exposed image, extracting structure and color information instead, and employs a multi-scale cross-attention to improve the feature fusion with long-exposed image. Second, as there is no existing large-scale training data for exposure fusion of dynamic scene, we also propose a novel training data synthesis pipeline, utilizing existing high-quality multi-exposure datasets (pre-aligned) of static scene and video datasets.
At last, to ensure the generated output maintains the fidelity to the reality, we also train an additional fidelity control branch, using the same decompose-and-fuse strategy and multi-scale cross-attention.


At last, to evaluate the effectiveness of our framework, we capture 100 under/over-exposed image pairs, covering daytime, nighttime, indoor, outdoor scenes with local and global motion patterns. Experiments on both latest HDR imaging datasets and our captured benchmark demonstrate that, comparing to existing methods, our \model is more robust to scenes with large exposure differences and large motion, as shown in~\cref{teaser}.

\section{Related work}
\label{sec:related}

\subsection{HDR imaging}

HDR Imaging can be devided into HDR reconstruction and Multi-Exposure Fusion (MEF) typically depends on the domain where fusion occurs~\cite{li2020fast}. HDR reconstruction methods~\cite{zheng2013hybrid, kalantari2017deep, wu2018deep, yan2019attention, yan2020deep, prabhakar2020towards, chen2022attention, liu2022ghost, song2022selective,tel2023alignment, xu2024hdrflow, kong2024safnet} inverts the camera response function (CRF) to merge exposure brackets in the linear HDR domain~\cite{ma2017robust}. In most cases, tone mapping is necessary to display the reconstructed HDR image properly on standard LDR monitors,.
As a cost-effective alternative~\cite{jiang2023meflut}, MEF methods~\cite{li2014selectively, zheng2015superpixel, li2017detail, mertens2007exposure, ram2017deepfuse, ma2017robust, prabhakar2019fast, li2020fast, xu2020u2fusion, liu2023emef, wu2024hsdsmef, jiang2023meflut, zhao2024film,zhu2024tcmoa} directly fuse images in the LDR domain, sidestepping CRF calibration and sophisticated tone mapping process~\cite{mertens2007exposure}.
Regardless of the type of HDR Imaging methods, they contend with ghosting artifacts caused by camera shake and object movement~\cite{shu2024RealHDRV}. Previous methods have attempted explicit or implicit alignment using optical flow~\cite{kalantari2017deep,prabhakar2019fast,prabhakar2020towards, wu2018deep} or attention mechanism~\cite{chen2022attention,liu2022ghost,song2022selective,tel2023alignment,yan2019attention,yan2020deep, kong2024safnet}. However, most HDR Imaging methods suffer from unpleasing artifacts when large motion causes occlusion in the complementary region.

\noindent\textbf{Diffusion models.} Recently, it has been witnessed with the rapid rise of diffusion models~\cite{ddpm, stablediffusion} and their successful application in various tasks, including controllable image generation~\cite{ye2023ip, controlnet}, image restoration~\cite{ozdenizci2023weatherdiffusion, lin2023diffbir, weng2024lcad, liu2024diffplugin, yu2024supir}, image editing~\cite{meng2021sdedit, chen2024anydoor, shi2024dragdiffusion, ju2024brushnet} and image inpainting~\cite{lugmayr2022repaint,corneanu2024latentpaint,xie2023smartbrush,yu2023inpaint}. In the field of HDR imaging, the application of diffusion models has primarily focused on HDR deghosting~\cite{yan2023diffhdr, guan2024hdrvdiff, hu2024hdrcvpr2024}. However, since these methods do not levearge the diffusion priors learned from large-scale datasets, their generalization ability is limited by the scale of the HDR dataset. While some recent works~\cite{li2024sagiri, goswami2024semantichdr} have employed diffusion priors, they tend to focus on single image HDR. Without another differently-exposed image as reference, the results generated by these methods lack sufficient reliability. Unlike previous methods, we utilize diffusion priors and use the short-exposed image as a reference to perform reliable inpainting in the highlight regions of the over-exposed image, thereby achieving natural and reliable HDR scene reconstruction. Compared to diffusion-based inpainting methods~\cite{lugmayr2022repaint, corneanu2024latentpaint, xie2023smartbrush, yu2023inpaint} that perform inpainting solely from scratch, we leverage information from short-exposure images to guide a more accurate inpainting process.

\noindent\textbf{Tone mapping methods.} The goal of tone mapping is to convert HDR images to LDR for display on standard screens while enhancing visual detail. Due to the challenge of obtaining ground-truth tone mapping results, unsupervised deep learning approaches have been developed using adversarial~\cite{vinker2021unpaired} and contrastive~\cite{cao2023unsupervised} learning. To address data limitations, \citet{cai2018SICE} manually curated training data by selecting the best results from 13 tone mapping methods for network learning~\cite{cai2018SICE,hu2022joint}. However, previous tone mapping algorithms, lacking robust image priors and facing data constraints, struggle with visually pleasing results and generalization in extreme high dynamic range scenes. By incorporating diffusion-based image priors, our method achieves aesthetic results even in challenging high dynamic range scenarios (see \cref{teaser}).

\begin{figure*}[!t]
	\centering
	\includegraphics[width=\linewidth]{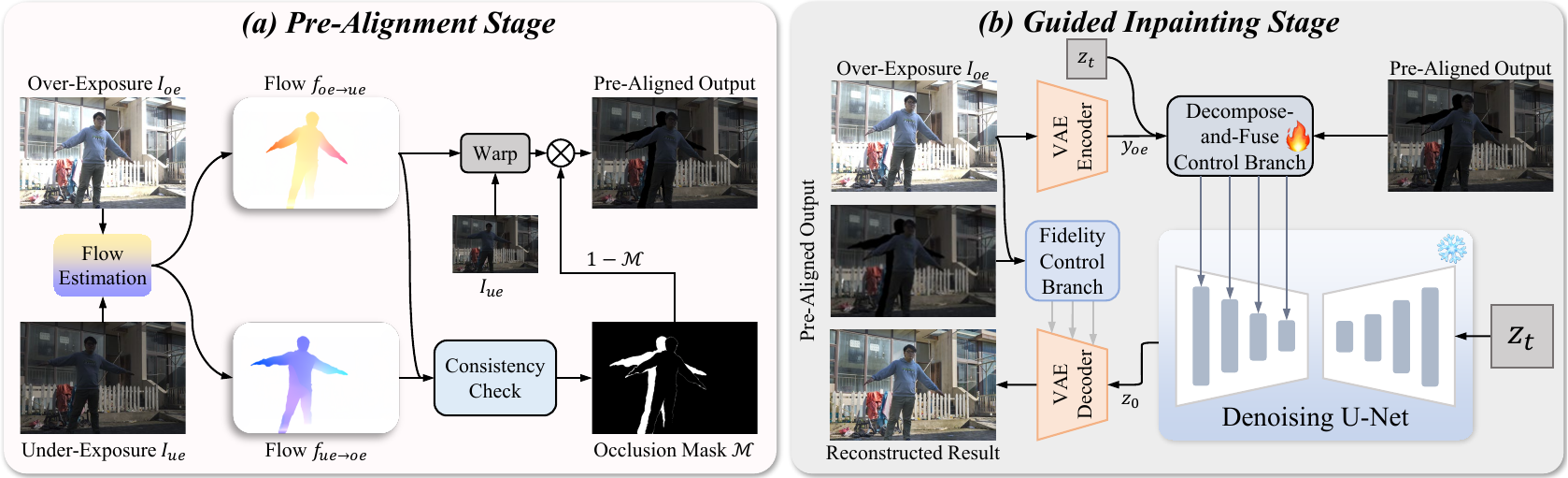}
    \vspace{-15pt}
	\caption{The whole backbone of \model. Our method is a 2-stage framework, consisting of (a) pre-alignment stage and (b) guided inpainting stage. The first stage pre-aligns the under-exposed image $I_{ue}$ to the over-exposed image $I_{oe}$ and masks the occluded regions. In the subsequent guided inpainting stage, we propose a new decompose-and-fuse control branch to utilize the diffusion priors, together with a fidelity control branch for fidelity keeping.}
	\label{fig:Backbone}
    \vspace{-15pt}
\end{figure*}

\section{Methodology}

Given an over-exposed image $I_{oe}$ and an under-exposed image $I_{ue}$, traditional exposure fusion algorithms directly aggregate different frequency band of both images, which are sensitive to misalignment error or lighting variation. Instead, we treat this as an inpainting problem. Specifically, we use the over-exposed image $I_{oe}$ as the base image and inpainting missing information in the highlight region. To ensure inpainted highlights are real, we also use highlights from under-exposed image as guidance.

Based on this idea, we design a 2-stage newtwork shown in~\cref{fig:Backbone}, which consists of the pre-alignment stage and the guided inpainting stage. The pre-alignment stage outputs a coarse-aligned version of $I_{ue}$, which is used as the soft guidance in the following guided inpainting stage. Details of each stage is described below.

\subsection{Pre-alignment stage}
Most of optical flow alignment assumes input have the similar brightness. Therefore, we first adjust the brightness $I_{ue}$ to match the distribution of $I_{oe}$ through intensity mapping function~\cite{grossberg2003IMF, liu2023unsupervised}. Then, we adopt RAFT~\cite{teed2020raft}, a pre-trained optical flow network, to estimate the bidirectional flow $f_{oe \rightarrow ue}$ and $f_{ue \rightarrow oe}$ and align $I_{ue}$ to $I_{oe}$ using backward warping. However, backward warping will result in ghosting at the occlusion boundary~\cite{zhao2020maskflownet}, leading to artifacts in the next guided inpainting stage. To solve that, we utilize the forward-and-backward consistency check~\cite{xu2022gmflow} to estimate occluded regions $\mathcal{M}$ and mask them out in the warped output. Finally, we can obtain a pre-aligned output $I_{ue \rightarrow oe}$ of the first stage:
\begin{equation}
    I_{ue \rightarrow oe} = (1 - \mathcal{M}) \cdot \mathcal{W}(I_{ue}, f_{oe \rightarrow ue}),
\end{equation}
where $\mathcal{W}$ denotes the backward warping.~\cref{fig:Backbone} (a) shows the output is a masked and aligned under-exposed image.

\subsection{Guided inpainting stage} 

We build our guided inpainting model based on the Stable Diffusion Model~\cite{Rombach_2022_CVPR} because its powerful generative prior can help to resolve ambiguity during inpainting. Similar to other diffusion-based image enhancement techniques~\cite{lin2023diffbir}, we also inject the following information through an additional control branch, as shown in~\cref{fig:Backbone}(b): 1) the image to be inpainted, which is the over-exposed image $I_{oe}$, 2) the additional guidance of highlight, which is the under-exposed image $I_{ue}$, and 3) the diffusion latent at the current diffusion step $z_t$, as previous work~\cite{lin2023diffbir} shows that including diffusion latent as an additional condition can improve image quality. The main diffusion denoising network is a pretrained U-Net. The over-exposed image is first encoded using a pretrained VAE before entering the diffusion module, and the outputs are converted back to the image space using the pretrained decoder.

The main differences between our solution and general diffusion-based image enhancement are two-fold. First, we propose a novel decompose-and-fuse control branch to inject two input images and diffusion latent as a control signal, as~\cref{fig:ablation} (b) shows that naively injecting this information may not be able to guide diffusion to faithfully inpaint missing highlights obtained from the under-exposed image. Second, we train an additional fidelity control branch to provide faithful structure and color information for the decoding process via shortcuts. Details are described below.


\begin{figure}[!t]
	\centering
	\includegraphics[width=\linewidth]{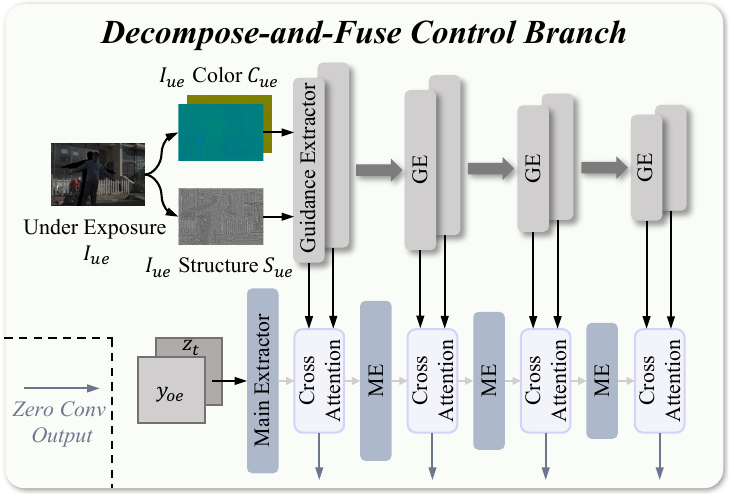}
	\caption{The detailed architecture of our proposed decompose-and-fuse control branch.}
    \vspace{-20pt}
	\label{fig:DFCB}
\end{figure}

\noindent\textbf{Decompose-and-fuse control branch.}
\cref{fig:DFCB} shows our control branch. We use the over-exposed image $I_{oe}$ as the main control signal and the under-exposed image $I_{ue}$ as the soft guidance. Following ControlNet~\cite{controlnet}, we copy the encoder and middle blocks of the denoising U-Net as the main extractor, but update their weights during training. A simple soft guidance is to use the encoded under-exposed image latent $y_{ue}$ from the VAE encoder, combined with the over-exposed latent $y_{oe}$. However, the underexposed images are often too dark to be directly used as soft guidance, since the model may entirely ignore this guidance.

Therefore, we decompose the under-exposed image into the color and structure information, both of which are robust to brightness changes. Specifically, we use the normalized image as the structure component, similar to SSIM~\cite{ssim}, as:
\begin{equation}
    S_{ue} = (Y_{ue} - \mu(Y_{ue})) / \sigma(Y_{ue}),
\end{equation}
where $Y_{ue}$ represents the luminance channel of $I_{ue}$ under YUV space, $\mu(\cdot)$ and $\sigma(\cdot)$ denote the mean intensity and standard deviation, respectively. The chroma channels (UV) are used as color information. The extracted structure and color information are further encoded using trained color and structure extractors (gray GE block in \cref{fig:DFCB}). Following~\cite{controlnet}, we implement GE using several simple convolution layers to extract multi-scale features.

At last, the extracted features are injected into the main extractor with a multi-scale cross attention, as shown in the bottom part of \cref{fig:DFCB}. The output of each cross attention module is fed into both the next level of the main extractor and the corresponding U-Net block, using zero convolution.

\noindent\textbf{Fidelity control branch.} Even with the control block, we sometimes observe undesirable modifications of texture introduced by VAE. An example is shown in~\cref{fig:ablation} (c). Therefore, to further improve fidelity, we design a Fidelity Control Branch (FCB), inspired by \cite{weng2024lcad}. FCB mitigates texture distortions by injecting features into the VAE decoder. It has a similar architecture as the decompose-and-fuse control branch, with two main differences: 1) the main extractor of FCB adopts the same structure as the VAE Encoder, rather than the denoising U-Net, to provide corresponding shortcuts to the VAE Decoder, with adjustments to the soft guidance extractor as well, and 2) the main extractor of the FCB directly takes the over-exposed image as input. To train the fidelity control branch, we freeze the pre-trained VAE Encoder and Decoder and encode the ground truth $I_{gt}$ to latent space, simulating the denoised latent $z_0$ during the inference. Then we input the corresponding over-exposed image and under-exposed image to the fidelity control branch to extract faithful features. At last, the VAE Decoder decodes the compressed latent to a reconstructed image $\hat{I}_{gt}$. We adopt the reconstruction loss $||\hat{I}_{gt} - I_{gt}||_1$ as an additional loss term.

\begin{figure}[t]
	\centering
	\includegraphics[width=\linewidth]{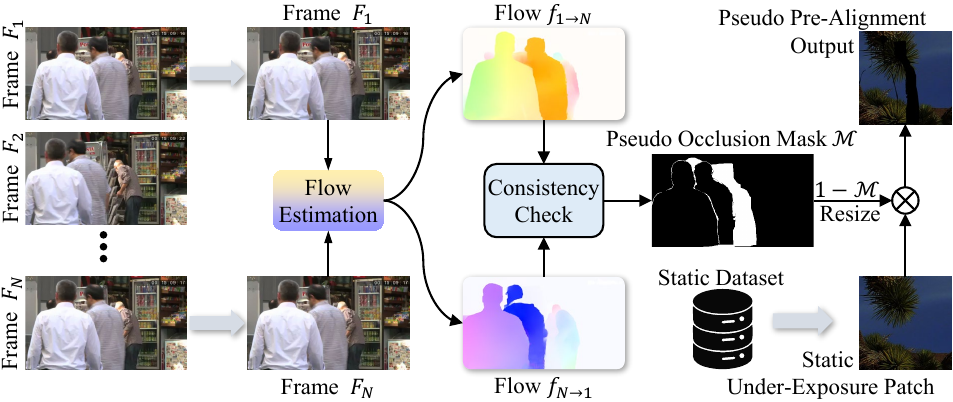}
    \vspace{-20pt}
	\caption{The illustration of our training data synthesis pipeline. 
    }
    \vspace{-15pt}
	\label{fig:datasyn}
\end{figure}

\subsection{Training data synthesis} Preparing the training data for the proposed guided inpainting network is challenging. To train the model, ideally we need a large-scale HDR dataset that 1)  covers different dynamic scenes, 2) has large exposure variance up to 9 stops, and 3) contains ground truth fusion result. However, no existing dataset satisfy all requirements. 


To solve this issue, we propose a novel training data synthesis pipeline. Specifically, as shown in~\cref{fig:datasyn}, we first randomly sample an image sequence with $N$ frames from a video dataset~\cite{xue2019Vimeo}. To model large motion, we select the first and the last frames. Then, similar to the pre-alignment stage, we estimate the bidirectional optical flow between two selected frames using~\cite{teed2020raft} and compute a pseudo occlusion mask via forward-backward consistency check. Subsequently, we randomly sample an under-exposed image patch from the high-quality static dataset~\cite{cai2018SICE} (ground truths are pre-aligned), resize the pseudo occlusion mask to match the patch size, and mask out the pseudo occluded region to synthesize a pseudo pre-aligned output. With our synthesized training data, our \model learns to handle dynamic scenes with only static multi-exposure image pairs.


\section{Experiment}

\subsection{Experimental setting}

\noindent\textbf{Datasets.} 
We utilize SICE~\cite{cai2018SICE} dataset and Vimeo-90K~\cite{xue2019Vimeo} dataset to synthesize our training data. Following previous works~\cite{cai2018SICE, liu2023emef}, we select images with the highest and lowest brightness in each exposure bracket of the SICE dataset as inputs. We evaluate our method on both static datasets and dynamic datasets. For evaluation, we use the MEFB dataset~\cite{mefb} with 100 static under/over-exposed image pairs and RealHDRV~\cite{shu2024RealHDRV}, a dynamic HDR deghosting test set containing 50 scenes with varying motion patterns.

\begin{figure}[!t]
	\centering
	\includegraphics[width=\linewidth]{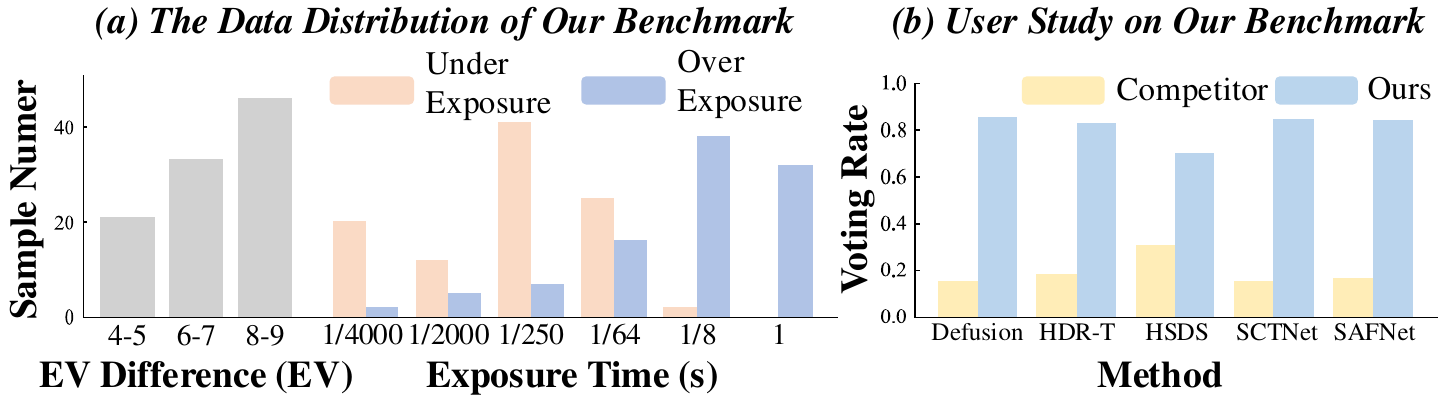}
    \vspace{-17pt}
	\caption{(a) The data distribution of our benchmark. Coordinate value of exposure time represents the upper boundary. (b) The user study result on our benchmark.}
    \vspace{-17pt}
	\label{fig:statistics}
\end{figure}

\begin{figure*}[!t]
    \centering
    \begin{minipage}{0.7\textwidth}
        \centering
        \captionof{table}{Quantitative comparisons on the static MEFB dataset~\cite{mefb}.}
        \adjustbox{width=\linewidth}{
        \setlength\tabcolsep{2pt}
        \fontsize{10}{10.5}\selectfont
		\begin{tabular}{l|c|ccccc}
			\toprule[1.2pt]
			\multirow{2}{*}{Type} & \multirow{2}{*}{Method} & \multicolumn{5}{c}{MEFB~\cite{mefb}} \\
			& & MUSIQ$\uparrow$ & DeQA-Score$\uparrow$ & PAQ2PIQ$\uparrow$ & HyperIQA$\uparrow$ & MEF-SSIM$\uparrow$\\
			\midrule \midrule
            \multirow{3}{*}{HDR Rec.} & HDR-Transformer~\cite{liu2022hdrtransformer} & 63.10 & 2.983 & 71.36 & 0.5996 & 0.8626 \\
            & SCTNet~\cite{tel2023sctnet} & 63.13 & 3.021 & 71.48 & 0.6068 & 0.8777 \\
            & SAFNet~\cite{kong2024safnet} & 61.70 & 2.518 & 72.67 & 0.5646 & 0.7711 \\
			\midrule
            \multirow{7}{*}{MEF} & Deepfuse~\cite{ram2017deepfuse}  & 52.58 & 2.945 & 67.96 & 0.4399 & 0.8968 \\
            & MEF-GAN~\cite{xu2020mefgan}  & 50.59 & 2.818 & 69.99 & 0.3745 & 0.7722\\
			& U2Fusion~\cite{xu2020u2fusion} & 63.39 & 3.219 & 72.23 & 0.5159 & 0.9304 \\
   		    & Defusion~\cite{liang2022defusion} & 62.70 & 3.118 & 72.82 & 0.5455 & 0.9062  \\
           	& MEFLUT~\cite{jiang2023meflut} & 65.71 & 3.277 & 71.21 & 0.5267 & 0.8608 \\
            & HSDS-MEF~\cite{wu2024hsdsmef} & 66.76 & 3.544 & 72.60 & 0.6026 & 0.9520 \\
            & TC-MoA~\cite{zhu2024tcmoa} & 64.60 & 3.355 &  71.85 & 0.5394 & \textbf{0.9636} \\
            \midrule
            Ours & UltraFusion & \textbf{68.82} & \textbf{3.881} & \textbf{73.80} & \textbf{0.6482} & 0.9385  \\
			
			\bottomrule[1.2pt]
		\end{tabular}
	   }  
        \label{tab:mefb}
    \end{minipage}
    \hfill
    \begin{minipage}{0.29\textwidth}
        \centering
        \includegraphics[width=0.9\linewidth]{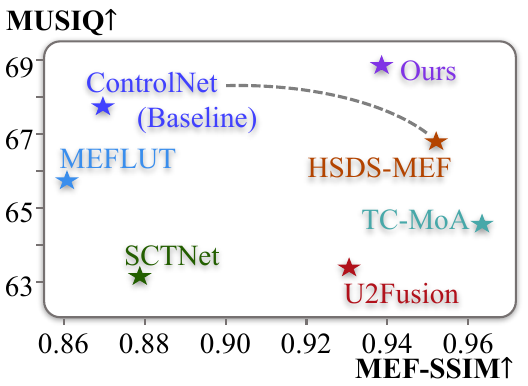}
        \caption{Trade-off curve between MEF-SSIM and MUSIQ on MEFB dataset~\cite{mefb}. Our \model~achieves the best trade-off between image quality and information preservation.}
        
        \label{fig:musiq}
    \end{minipage}%
\end{figure*}

\begin{table*}[h]
	\centering
    \vspace{-5pt}
	\caption{Quantitative comparisons on dynamic RealHDRV dataset~\cite{shu2024RealHDRV} and our challenging UltraFusion benchmark.}
    \vspace{-5pt}
	\label{tab:RealHDRV}
	\adjustbox{width=\linewidth}{
		\begin{tabular}{l|c|ccccc|cccc}
			\toprule[1.2pt]
			\multirow{2}{*}{Type} & \multirow{2}{*}{Method} & \multicolumn{5}{c|}{RealHDRV~\cite{shu2024RealHDRV}} & \multicolumn{4}{c}{UltraFusion Benchmark} \\
			& & TMQI$\uparrow$ & MUSIQ$\uparrow$ & DeQA-Score$\uparrow$ & PAQ2PIQ$\uparrow$ & HyperIQA$\uparrow$ & MUSIQ$\uparrow$ & DeQA-Score$\uparrow$ & PAQ2PIQ$\uparrow$ & HyperIQA$\uparrow$\\
			\midrule \midrule
            \multirow{3}{*}{HDR Rec.} & HDR-Transformer~\cite{liu2022hdrtransformer} & 0.8680 & 62.24 & 3.496 & 70.33 & 0.5225 & 63.66 & 2.909 & 72.83 & 0.5619  \\
            & SCTNet~\cite{tel2023sctnet} & 0.8715  & 62.69 & 3.532 & 70.74 & 0.5272 & 61.84 & 3.102 & 72.94 & 0.5888 \\
            & SAFNet~\cite{kong2024safnet} & 0.8726 & 62.07 & 3.506 & 70.48 & 0.5156 & 61.50 & 2.179 & 73.15 & 0.5487 \\
			\midrule
            \multirow{3}{*}{MEF} & Defusion~\cite{liang2022defusion} & 0.8187 & 56.60 & 3.302 & 68.38 & 0.4856 & 60.31 & 3.352 & 71.87 & 0.5463 \\
            & MEFLUT~\cite{jiang2023meflut} & 0.8297 & 62.42 & 3.315 & 70.04 & 0.5020 & 63.62 & 3.343 & 71.73 & 0.5074 \\
            & HSDS-MEF~\cite{wu2024hsdsmef} & 0.8323 & 61.76 & 3.360 & 71.11 & 0.5054 & 64.54 & 3.627 & 73.42 & 0.5923 \\
            \midrule
            Ours & UltraFusion & \textbf{0.8925} & \textbf{67.51} & \textbf{3.830} & \textbf{73.40} & \textbf{0.5833} & \textbf{68.41} & \textbf{3.957} & \textbf{75.18} &\textbf{0.6214}  \\
			
			\bottomrule[1.2pt]
		\end{tabular}
	}
    \vspace{-5pt}
\end{table*}

\begin{figure*}[!t]
	\centering
	\includegraphics[width=\linewidth]{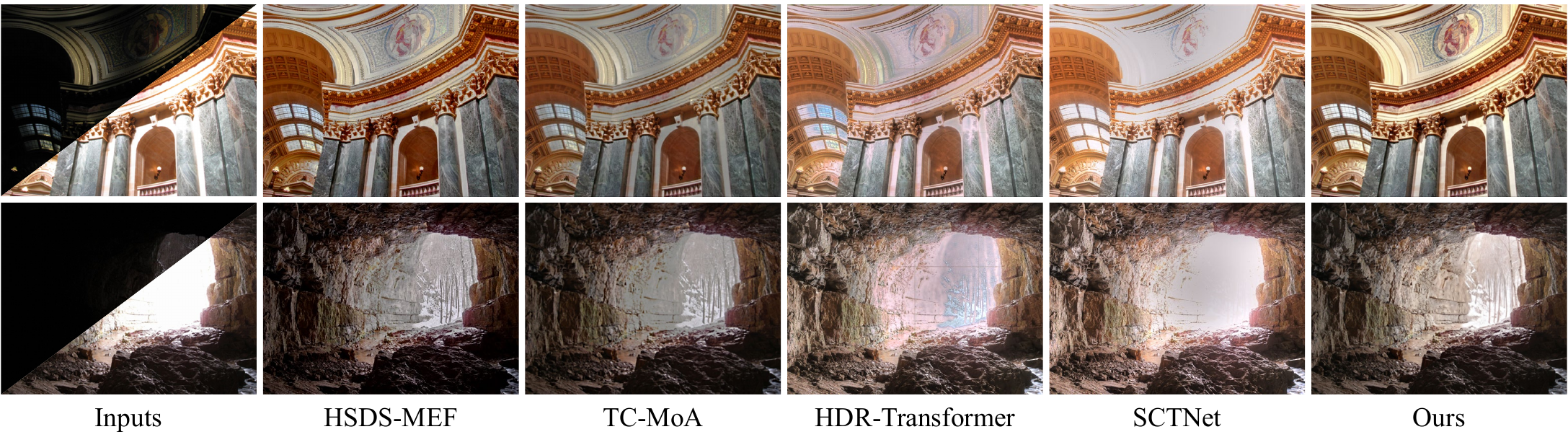}
    \vspace{-20pt}
	\caption{Visual comparisons of different exposure fusion methods on static MEFB dataset~\cite{mefb}. 
    }
    \vspace{-15pt}
    \label{fig:MEFB}
\end{figure*}

\noindent\textbf{UltraFusion benchmark.}
Existing exposure fusion benchmark cannot fully evaluate fusing in real-world challenging conditions, as they either lack realistic motion~\cite{mefb} or have limited dynamic range~\cite{kalantari2017deep, tel2023sctnet, shu2024RealHDRV}. Therefore, we collect a new real-world UltraFusion benchmark, which contains 100 real-captured under/over-exposed image pairs. Compared to previous datasets, our benchmark is more challenging for three reasons: 1) Our benchmark features larger exposure differences between the two input images (up to 9 stops). 2) It includes more realistic motion, with many scenes containing extensive and unintentional foreground movement. 3) Our benchmark is highly diverse, encompassing daytime, nighttime, indoor, and outdoor scenes captured by DSLR Camera (Canon R8) and mobile phones (iPhone12, iPhone13, Redmi K50 Pro and OPPO Reno8 Pro). We summarize the exposure difference distribution and exposure time distribution of our benchmark in~\cref{fig:statistics} (a). It can be observed that our benchmark covers a wide range of exposure differences and diverse exposure times, which can be used to effectively test the robustness of the HDR methods.

\noindent\textbf{Implementation details.}
We leverage the generative prior encapsulated in Stable Diffusion V2.1~\cite{stablediffusion}. The decompose-and-fuse control branch (DFCB) is trained for 140k iterations with batch size $bs = 32$ on 8 NVIDIA RTX 4090 GPUs. We also train the fidelity control branch for 1000K iteration with batch size $bs = 1$ on a single NVIDIA RTX 4090 GPU. Adam is adopted as the optimizor and the learning rate is fixed to 0.0001. To adapt HDR reconstruction methods to 2 differently exposed inputs, we re-implement them by following their default settings. 

\noindent\textbf{Evaluation metrics.}
We utilize four widely used non-reference metric MUSIQ~\cite{musiq}, DeQA-Score~\cite{deqascore}, PAQ2PIQ~\cite{paq2piq} and HyperIQA~\cite{hyperiqa} for quantitative comparison.
Moreover, for static dataset~\cite{mefb}, as no ground truths are provided, we select the task-specific MEF-SSIM~\cite{mefssim} for structure retention evaluation. For dynamic dataset~\cite{shu2024RealHDRV} with HDR ground truths, TMQI~\cite{tmqi} is used to evaluate the performance of fidelity and naturalness. In addition, we conduct a user study on our Ultrafusion benchmark to perform subjective evaluation.

\subsection{Comparisons with HDR imaging methods}
We compare our method with state-of-the-art HDR Imaging methods, including HDR reconstruction methods HDR-Transformer~\cite{liu2022hdrtransformer}, SCTNet~\cite{tel2023sctnet}, SAFNet~\cite{kong2024safnet}, and multi-exposure fusion methods Deepfuse~\cite{ram2017deepfuse}, MEF-GAN~\cite{xu2020mefgan}, U2Fusion~\cite{xu2020u2fusion}, Defusion~\cite{liang2022defusion}, MEFLUT~\cite{jiang2023meflut}, HSDS-MEF~\cite{wu2024hsdsmef}, TC-MoA~\cite{zhu2024tcmoa}. As HDR reconstruction methods cannot output an LDR image directly, we use professional software Photomatix~\cite{Photomatix} to perform tone mapping.

\begin{figure*}[!t]
	\centering
	\includegraphics[width=\linewidth]{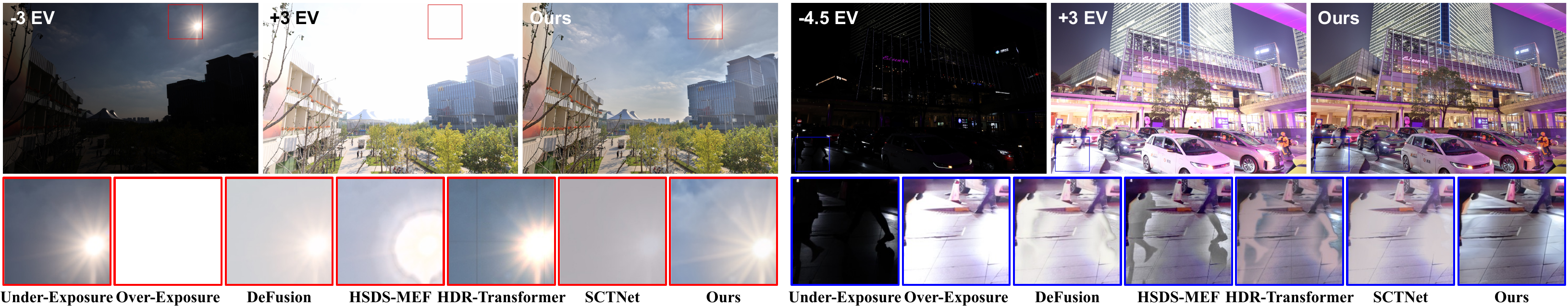}
    \vspace{-17pt}
	\caption{Visual comparisons on our captured UltraFusion benchmark.}
    \vspace{-15pt}
	\label{fig:OurBenchmark}
\end{figure*}

\noindent\textbf{Evaluation on static dataset.} 
We evaluate the fusion performance on the MEFB dataset~\cite{mefb}, focusing on large exposure differences. As shown in~\cref{tab:mefb}, our method outperforms other methods on all four non-reference metrics (MUSIQ, DeQA-Score, PAQ2PIQ, and HyperIQA). Specifically, our proposed UltraFusion achieves 2.06 gain in terms of MUSIQ compared to HSDS-MEF. 
In terms of MEF-SSIM, as shown in ~\cref{fig:musiq}, our baseline model (ControlNet~\cite{controlnet}) achieves better image quality but lacks fidelity, while HSDS-MEF~\cite{wu2024hsdsmef} retains more information from inputs at the cost of quality. Our \model outperforms most algorithms, and achieves similar fidelity scores compared to HSDS-MEF and TC-MoA but with much higher image quality (non-reference metrics), indicating the best trade-off between visual quality and information preservation.  The qualitative comparison in~\cref{fig:MEFB} further validates this. In contrast, HDR reconstruction methods (HDR-Transformer and SCTNet) often miss some detail in highlights and MEF methods (HSDS-MEF and TC-MoA) introduce unnatural transition from brighter to dark regions.

\begin{figure}[!t]
    \centering
    \includegraphics[width=\linewidth]{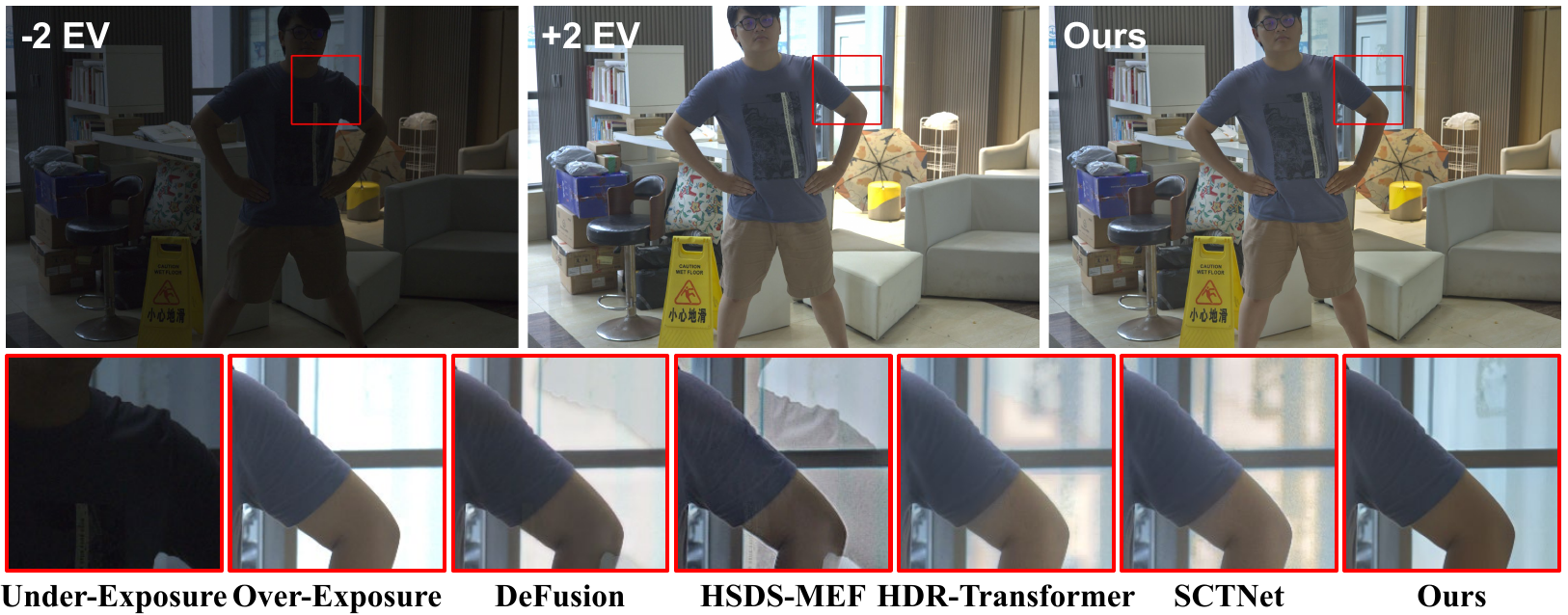}
    \vspace{-17pt}
    \caption{Visual results on dynamic RealHDRV dataset~\cite{shu2024RealHDRV}.}
    \vspace{-20pt}
    \label{fig:RealHDRV}
\end{figure}

\noindent\textbf{Evaluation on dynamic dataset.} To further illustrate the robustness of \model to global and local motion, we use the RealHDRV dataset~\cite{shu2024RealHDRV}. We extract the corresponding over-exposed image from the HDR ground truth as input. ~\cref{tab:RealHDRV} demonstrates that our UltraFusion achieve state-of-the-art performance in terms of all metrics. For dynamic scenes, TMQI metric is a particularly important metric, as it is specially designed for HDR evaluation by assessing structural similarity between fusion output and ground truth. Since MEF methods are mainly designed for static scenes, they lack the capability to handle motion, resulting in low TMQI scores and overlay artifacts, as shown in~\cref{fig:RealHDRV}. HDR reconstruction methods trained on dynamic datasets achieve better performance, but still produce noticeable artifact. On the other side, due to our soft inpainting guidance, \model is much more robust to misalignment and occlusion and the fusion output contains almost no artifacts. It achieves the highest TMQI in \cref{tab:RealHDRV} and the best visual result in~\cref{fig:RealHDRV}.

\noindent\textbf{Evaluation on our UltraFusion benchmark.}
At last, the evaluation on our benchmark validates the robustness of \model in the most challenging scenes. On all four non-reference metrics, our method outperforms compititors by a large margin, as shown in~\cref{tab:RealHDRV}. The qualitative comparison in~\cref{fig:OurBenchmark} is also consistent with the quantitative metrics. 
For example, in the red zoom-in patch, integrating the highly bright sun from the under-exposed image into the over-exposed image is extremely challenging. Other methods fail to maintain the shape of the sun or preserve high contrast in the fused region, while our method naturally reconstructs the sun, preserving its appearance and visual-pleasing tone of the whole image.
Moreover, we conduct a user study on our proposed benchmark. Specifically, we randomly select 20 scenes from our benchmark and invite 136 different users to participate. For each scene, each user is asked to compare our method with a randomly chosen baseline. The user study in~\cref{fig:statistics} (b) indicates our method is more favored by users than competitors. This outcome aligns with the non-reference metric evaluation, showing that our method produces more natural images. \textbf{More visual comparisons are available in the supplementary}.

\subsection{Ablation studies}

\begin{table}[t]
    \centering
    \caption{Ablation studies of three key components of our proposed \model on RealHDRV dataset~\cite{shu2024RealHDRV}.}
    \vspace{-7pt}
	\label{tab:ablation}
	\adjustbox{width=\linewidth}{
		\begin{tabular}{c|cc}
			\toprule[1.2pt]
			Model & TMQI$\uparrow$ & MUSIQ$\uparrow$ \\
			\midrule 
            w/o Alignment Strategy & 0.7427 & 63.67 \\
            w/o Decompose-and-Fuse Control Branch & 0.8872 & 66.94 \\
            w/o Fidelity Control Branch & 0.8763 & 67.36 \\
            Ours UltraFusion & \textbf{0.8925} & \textbf{67.51}\\
			\bottomrule[1.2pt]
		\end{tabular}
	}
    \vspace{-15pt}
\end{table}

To validate the effectiveness of our UltraFusion, we conduct ablation studies on the three key components, followed by an in-depth exploration of the designs of them.

\begin{figure}[t]
	\centering
	\includegraphics[width=\linewidth]{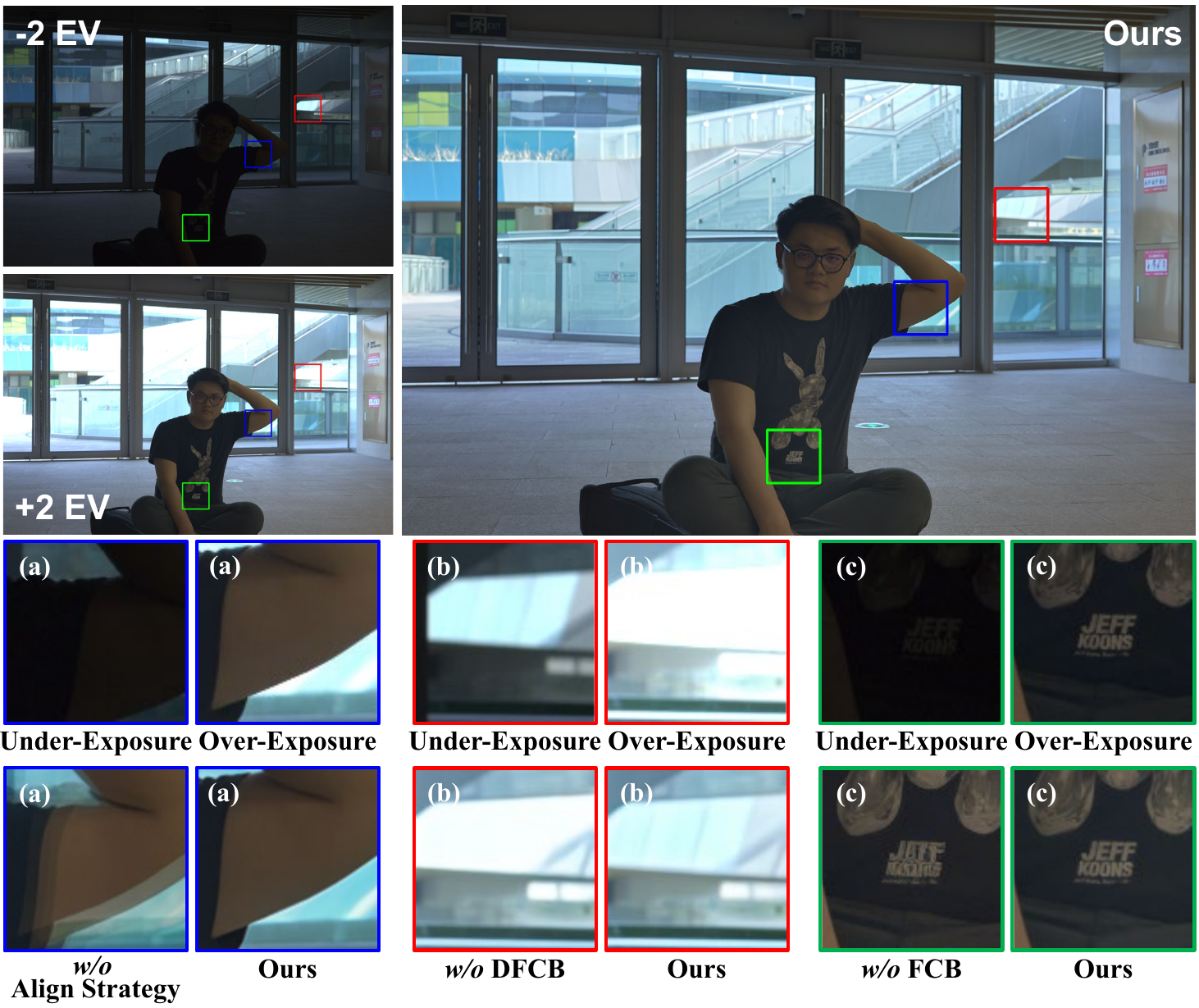}
    \vspace{-17pt}
	\caption{Visual results of ablating each key component of our method. Each key component contributes to the final results.}
    \vspace{-8pt}
	\label{fig:ablation}
\end{figure}

\noindent\textbf{Key components.}
First, we perform the ablation study of the proposed three key components, including the alignment strategy, decompose-and-fuse control branch, and fidelity control branch, on the RealHDRV~\cite{shu2024RealHDRV} dataset, as shown in \cref{tab:ablation}. We first remove our alignment strategy, opting instead to train the model directly on the original SICE~\cite{cai2018SICE} dataset and input differently exposed image pairs without coarse alignment. Without the pre-alignment stage and the data synthesis pipeline designed for it, performance drops significantly in terms of TMQI. \cref{fig:ablation} (a) also shows that the model fails to implicitly align large motions. Then, we replace the decompose-and-fuse control branch (DFCB) with the vanilla ControlNet~\cite{controlnet}, and it fails to fuse features with large exposure differences, which leads to the loss of details (see ~\cref{fig:ablation} (b)) and a decrease in TMQI. Finally, when we exclude the fidelity control branch (FCB), \model loses the capability to maintain the detailed structure and vivid color, as indicated by~\cref{fig:ablation} (c).

\begin{figure}[t]
    \centering
    \includegraphics[width=\linewidth]{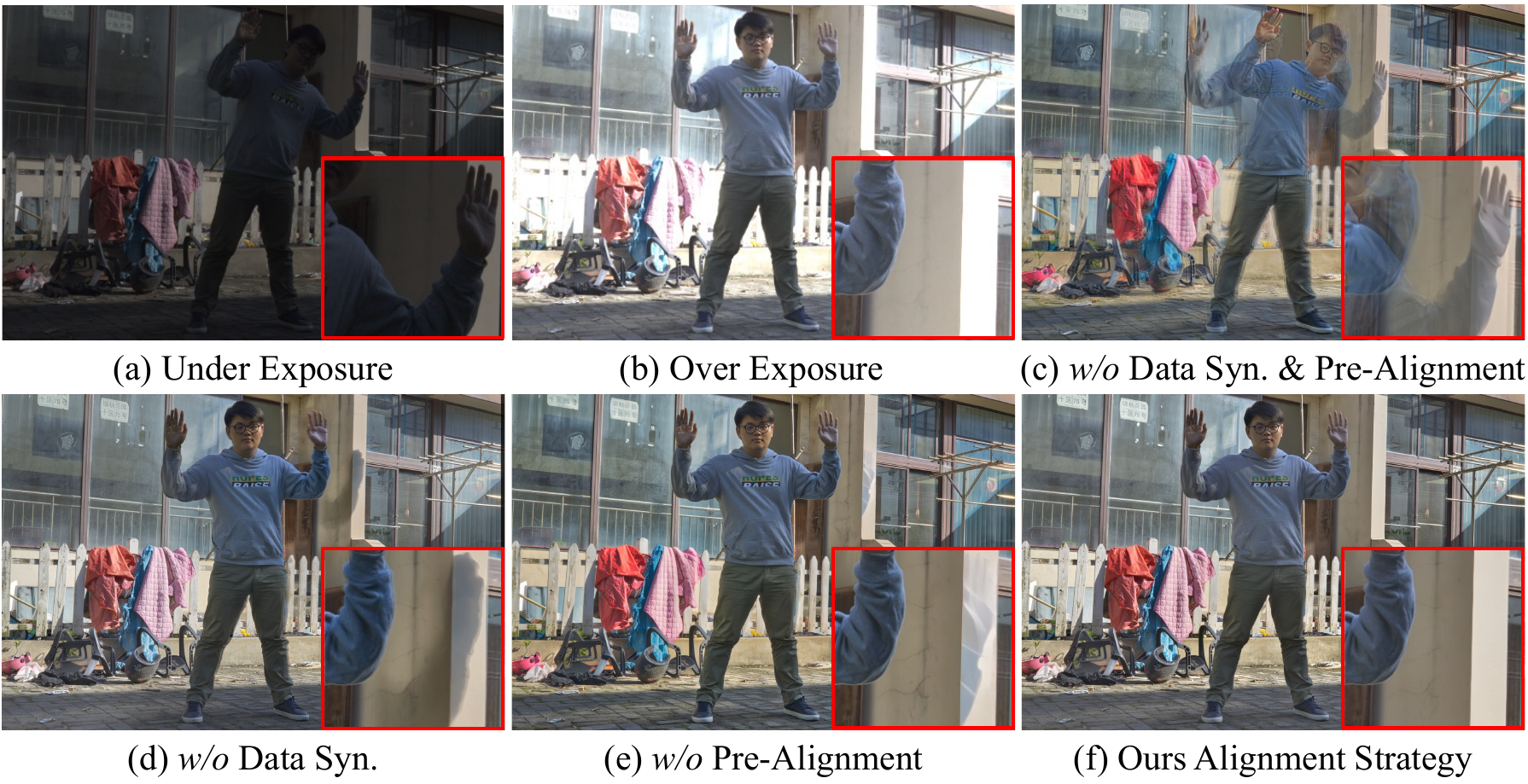}
    \vspace{-19pt}
    \caption{Effectiveness of our alignment strategy.}
    \vspace{-18pt}
    \label{fig:ablation_framework}
\end{figure}

\noindent\textbf{Alignment strategy.}
Second, we perform ablation of the proposed alignment strategy, which is the most critical design for our framework. Our alignment strategy mainly consists of a data synthesis pipeline at training time and a pre-alignment module at testing time.
Without both of them, the whole model degrades to a multi-exposure fusion method as shown in \cref{fig:ablation_framework} (c). Performing pre-alignment can reduce artifacts to some extent (see \cref{fig:ablation_framework} (d)), but may still be prone to alignment errors. Our data synthesis pipeline can improves the robustness of our algorithm to unaligned conditions, but motion in the occluded regions still cannot be solved, as shown in~\cref{fig:ablation_framework}. After combining data synthesis pipeline and pre-alignment stage, our algorithm demonstrates strong capability towards dynamic scenes.

\begin{figure}[t]
    \centering
    \includegraphics[width=\linewidth]{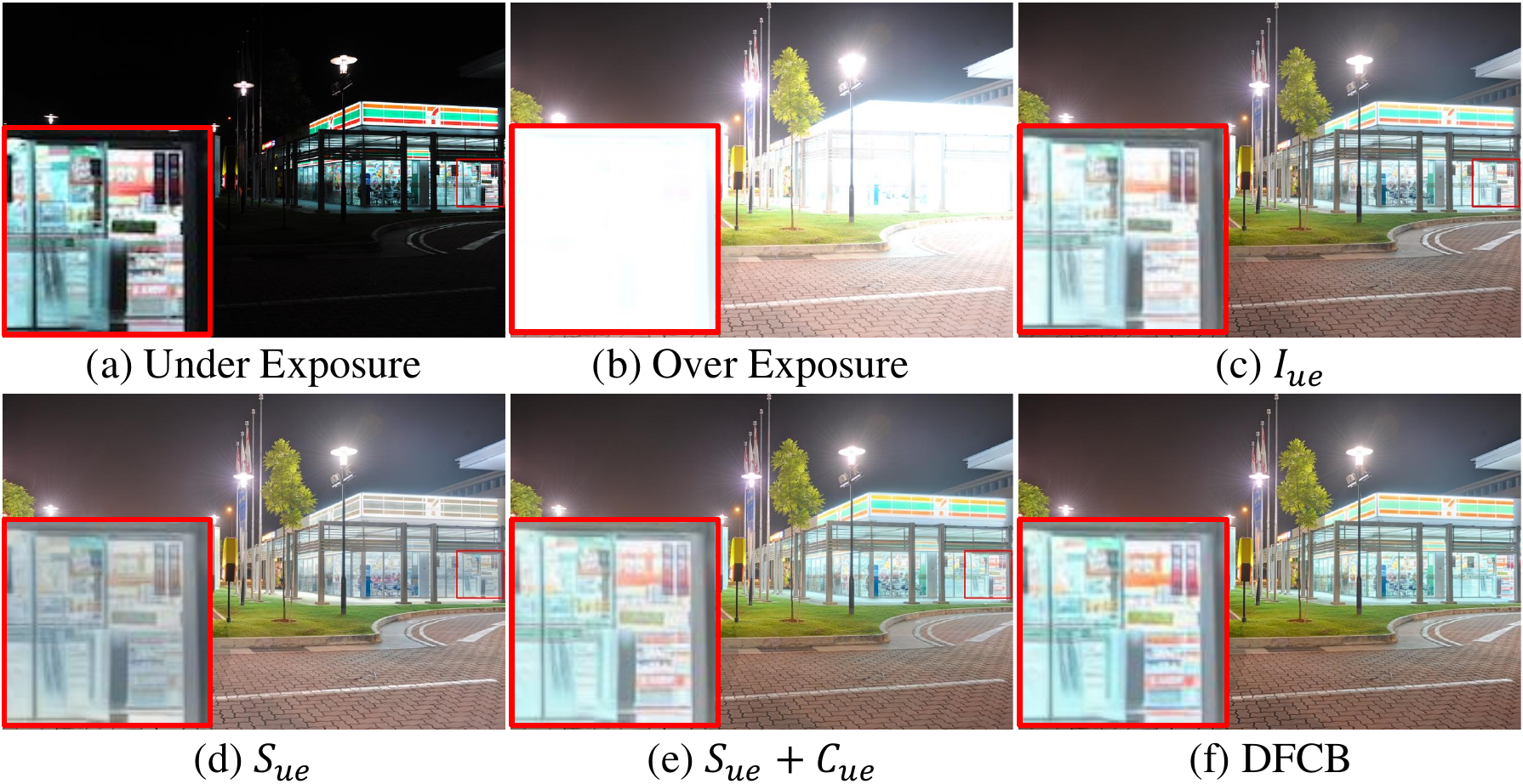}
    \vspace{-17pt}
    \caption{Detailed ablation study on the design choices of decompose-and-fuse control branch.}
    \vspace{-8pt}
    \label{fig:ablation_dfcb}
\end{figure}

\noindent\textbf{Decompose-and-fuse control branch.}
At last, we evaluate how the form of soft guidance provided by the under-exposed image impacts the recovery of the highlight regions. When using the RGB under-exposed image $I_{ue}$ as guidance, the model will ignore some details in the output (\cref{fig:ablation_dfcb} (c)). Replacing $I_{ue}$ with its structure information $S_{ue}$ retains more details (\cref{fig:ablation_dfcb} (d)), but without generating vivid color. By incorporating both under-exposed color information $C_{ue}$ and structure information $S_{ue}$, the reconstructed results can maintain more details and color consistency (see \cref{fig:ablation_dfcb} (e)). At last, multi-scale cross-attention further improves the fusion result (see \cref{fig:ablation_dfcb} (f)).



\subsection{Application on general image fusion}
\begin{figure}[t]
    \centering
    \includegraphics[width=\linewidth]{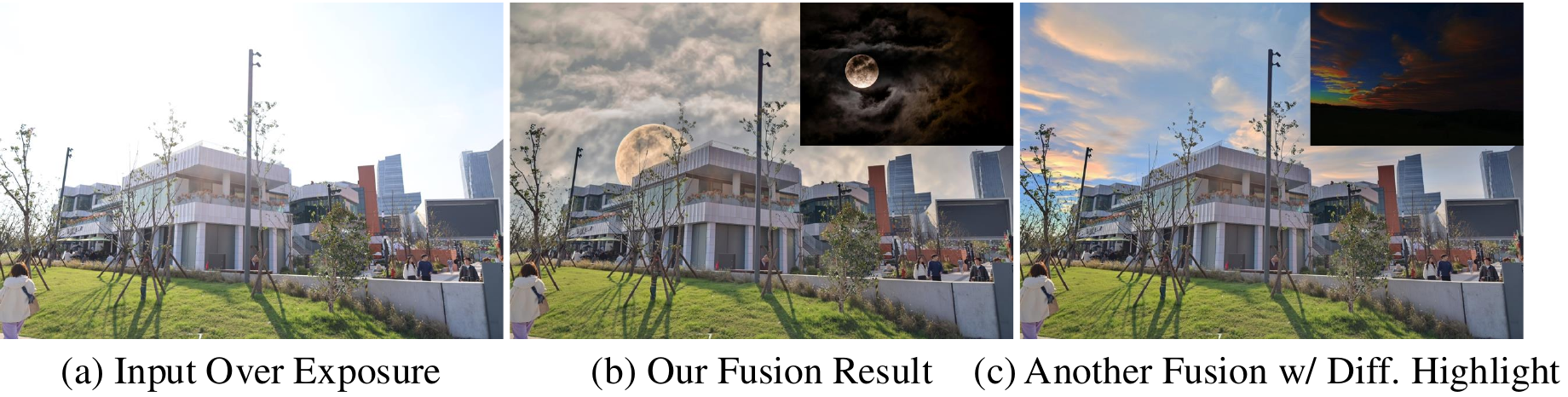}
    \vspace{-19pt}
    \caption{Extension to general fusion. Given totally different underexposed images as guidance (upper right corner), our method can also generate different fusion results. 
    }
    \vspace{-16pt}
    \label{fig:extension}
\end{figure}
One advantage of the \model is that it can be extended to general image fusion, thanks to the flexibility of our guided inpainting. To illustrate this potential, we explore one additional interesting demo to fuse two irrelevant images captured by different cameras at different locations. As shown in~\cref{fig:extension}, \model successfully copies the moon (b) or blue sky (c) to the over-exposed image, unlocking many interesting potential applications (\eg, image harmonization).

\vspace{-5pt}
\section{Conclusion}
\vspace{-5pt}
In this work, we introduce a novel approach to HDR imaging, tackling challenges presented by significant exposure differences and large motion. By modeling the fusion process as a guided inpainting problem and utilizing the under-exposed image for soft guidance, our method is robust with alignment errors and circumvents tone mapping, resulting in natural, artifact-free outputs. We also propose decompose-and-fuse control branch and fidelity control branch to improve feature modulation and fidelity preservation of ControlNet. Extensive experiments on existing datasets and our captured benchmark demonstrate the robustness and effectiveness of our method compared to previous HDR methods.

With extreme exposure differences and challenging non-rigid motion, occlusion mask estimation may introduce errors, causing the restoration of certain highlight regions degrades to single image HDR. While our method is able to use diffusion priors to restore these highlights, restoration without under-exposed information remains unreliable. Moreover, it takes almost 3.3s to fuse $512 \times 512$ inputs on an NVIDIA RTX 4090 GPU. A more exposure-robust optical flow algorithm and a faster implementation are highly desirable, and we leave them for future work.

\section*{Acknowledgment}
This work was supported by the National Key R\&D Program of China No.2022ZD0160201, Shanghai Artificial Intelligence Laboratory and RGC Early Career Scheme (ECS) No. 24209224.

{
    \small
    \bibliographystyle{ieeenat_fullname}
    \bibliography{main}
}

\appendix
\renewcommand\thefigure{A\arabic{figure}}
\renewcommand\thetable{A\arabic{table}}  
\renewcommand\theequation{A\arabic{equation}}
\setcounter{section}{0}
\setcounter{equation}{0}
\setcounter{table}{0}
\setcounter{figure}{0}

\clearpage
\setcounter{page}{1}
\maketitlesupplementary

\section{Why We Need Handle 9-Stops?}
\begin{figure}[t]
	\centering
	\includegraphics[width=\linewidth]{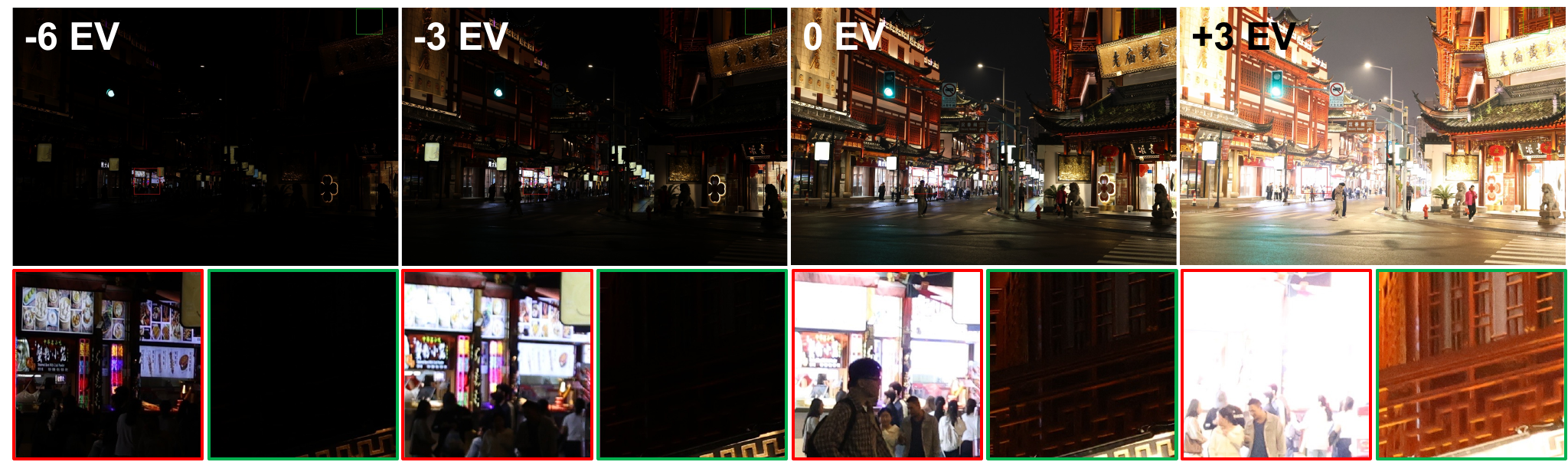}
    \caption{An example of 9-stops scenes.}
    \label{fig:9stops}
\end{figure}

Some challenging night-time scenes require up to 9 stops of exposure difference to cover the full dynamic range. As shown in~\cref{fig:9stops}, we need -6 EV to capture highlights (red box) and +3 EV (green box) to capture dark details.

\begin{figure}[t]
	\centering
	\includegraphics[width=\linewidth]{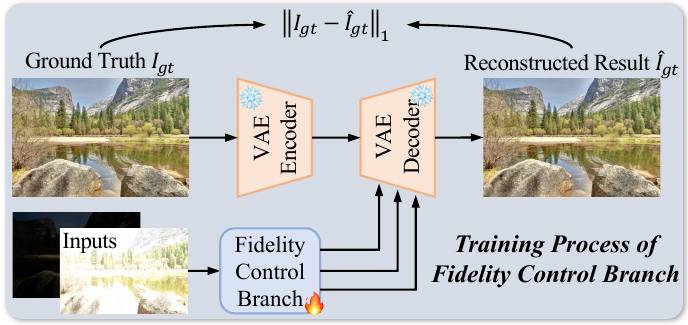}
    \caption{Detailed process of training FCB.}
    \label{fig:fcb_train}
\end{figure}

\section{Training process of Fidelity Control Branch}
To better illustrate how fidelity control branch is implemented, we show its training process in~\cref{fig:fcb_train}. Unlike the inference stage of our UltraFusion, the input of the VAE during FCB training is the ground truth. Our goal is to enable FCB to learn features that assist VAE decoding through shortcuts.

\section{Evluation Details}
In RealHDRV~\cite{shu2024RealHDRV} dataset, the HDR ground truth corresponds to the 0 EV input.
However, many 0 EV images in RealHDRV~\cite{shu2024RealHDRV} dataset only contain few over-exposed regions need to be recovered. To better demonstrate ultra high dynamic imaging performance of various methods, we extract LDR image of 2 EV or 3 EV (according to the under-exposed input is -2 EV or -3 EV) from HDR groundtruth as over-exposed input, by reversing the process adopted to fuse the HDR groundtruth. Finally, after our augmentation, the RealHDRV~\cite{shu2024RealHDRV} dataset contains 50 paired under/oever-exposed inputs with 4 or 6 stops.

\section{Ablation Study on Fidelity Control Branch}
\begin{figure}[t]
	\centering
	\includegraphics[width=\linewidth]{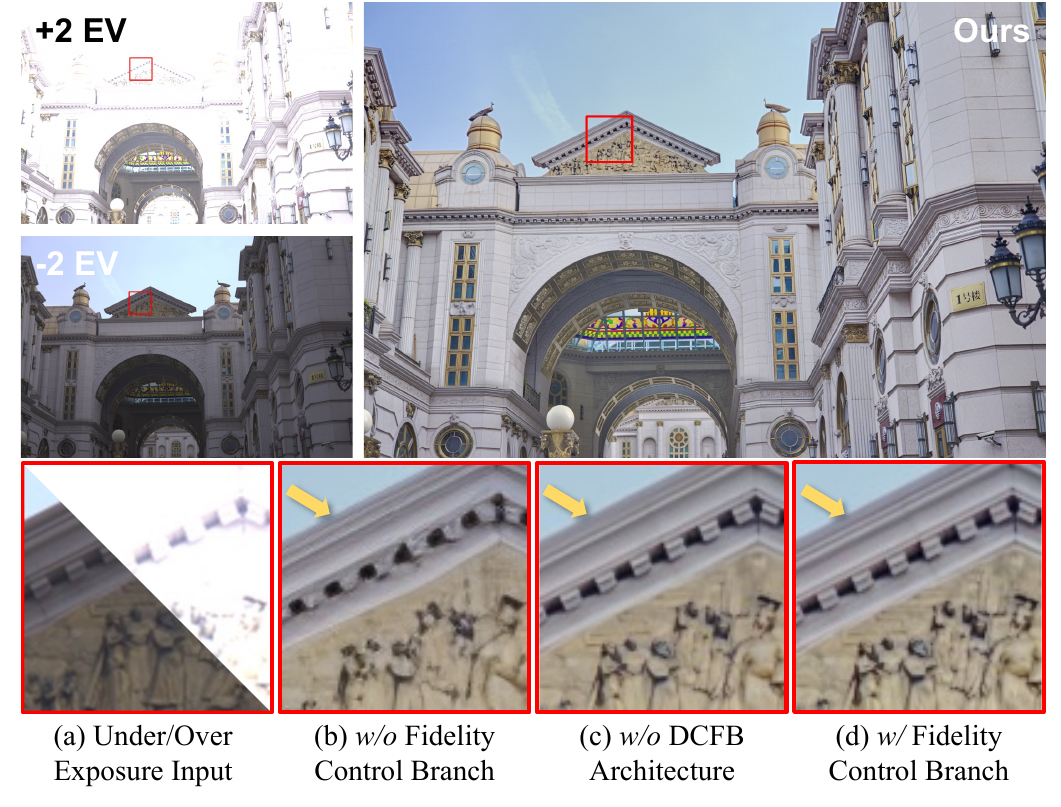}
	\caption{Illustrating the effectiveness of leveraging the similar architecture as decompose-and-fuse control branch in fidelity control branch.}
	\label{fig:ablation_fe}
\end{figure}

As shown in \cref{fig:ablation_fe}, the fidelity control branch effectively preserves the faithful structure of inputs. However, simply using two RGB images as inputs leads to some texture loss, as shown in \cref{fig:ablation_fe}(c). We demonstrate in \cref{fig:ablation_fe}(d) that by adopting similar architecture as decompose-and-fuse control branch (DCFB), more high-frequency details are retained and the overall visual quality is enhanced.

\section{Cross Attention Architecture}
\begin{figure}[t]
	\centering
	\includegraphics[width=\linewidth]{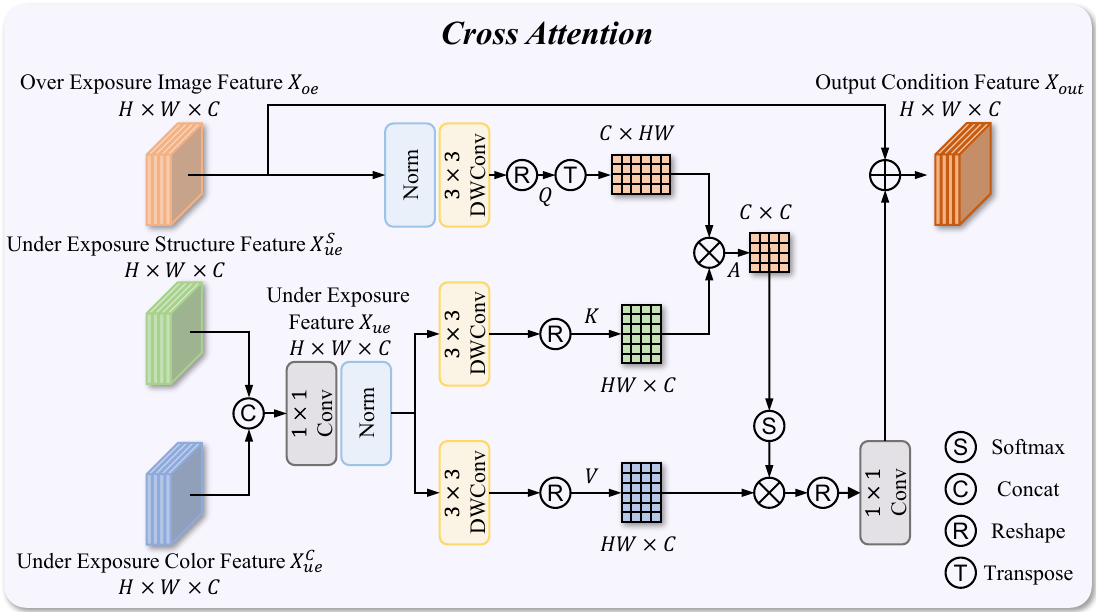}
	\caption{Detailed architecture of cross attention.}
	\label{fig:cross_attn}
\end{figure}

We utilize cross attention in the decompose-and-fuse control branch to fuse features from different modalities. The structure of the cross attention module is illustrated in ~\cref{fig:cross_attn}. The cross attention module accepts three inputs, \ie, overexposed image feature $X_{oe} \in \mathbb{R}^{H\times W\times C}$, short-exposed structural features $X_{ue}^S \in \mathbb{R}^{H\times W\times C}$, and short-exposed color features $X_{ue}^C \in \mathbb{R}^{H\times W\times C}$. First, we concatenate $X_{ue}^S$ and $X_{ue}^C$ and apply an $1 \times 1$ convolution to adjust channel dimension back to $C$, obtaining the under exposure feature $X_{ue}$. Subsequently, LayerNorm is separately applied to $X_{oe}$ and $X_{ue}$, followed by $3 \times 3$ depth-wise convolutions to produce the corresponding $Q$, $K$ and $V$.
Next, we perform attention operations on obtained $Q$, $K$ and $V$. After reshaping the output of attention operation, we input it to another $1 \times 1$ convolution layer and add the result to $X_{oe}$ to produce output condition feature $X_{out}$. The whole process can be summarized as follows:
\begin{equation}
    X_{out} =X_{oe} + \text{Conv}_{1\times1}(V\text{Softmax}(\frac{Q^TK}{\tau})), 
\end{equation}
where $\tau$ is a learnable scaling factor.

\section{Extend to 3 Exposures}
\begin{figure}[t]
	\centering
	\includegraphics[width=\linewidth]{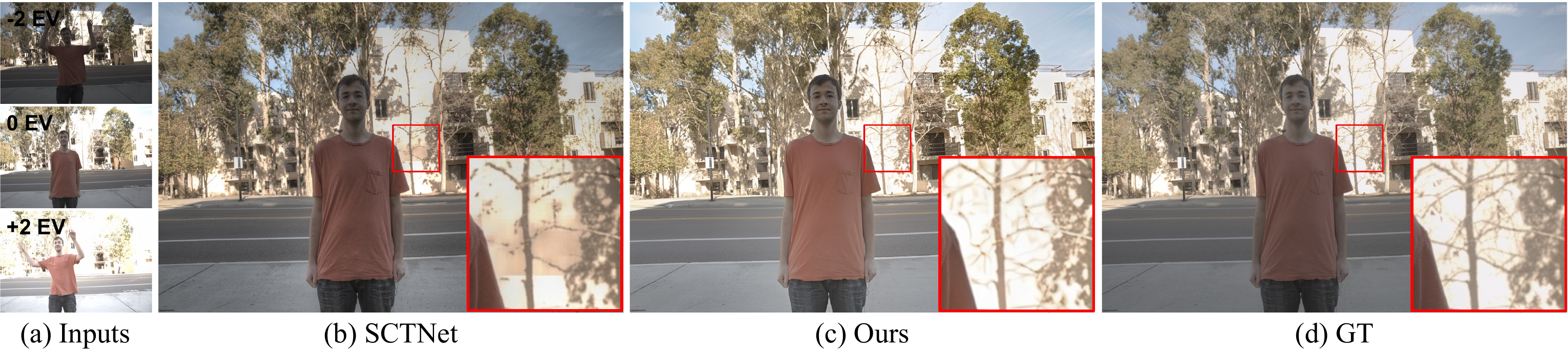}
    \caption{Visual comparison with SCTNet~\cite{tel2023sctnet} on Kalantari's dataset~\cite{kalantari2017deep}. Our framework can be extended to 3 exposures flexibly.}
    \label{fig:kalantari}
\end{figure}

Our UltraFusion is focus on 2 exposures as it already generates very good results and reduces the user's capture burden. Extending to 3 exposures is straightforward. We use the normal-exposed image as the reference and process it similarly. For the other two exposures, we extract guided features using the guidance extractor, then use normalized summation of them as input to the cross attention module. In the 3-exposure setup, we train UltraFusion on Kalantari's dataset~\cite{kalantari2017deep} according to conventional settings and test on the corresponding test set. The comparison is performed with officially released state-of-the-art HDR reconstruction methods. The qualitative results are shown in~\cref{fig:kalantari}, respectively.

\section{Effectiveness of Pre-Alignment}
To conduct a more fair comparison, we also pre-align the test set and summarize the performance of each competing method in~\cref{tab:RealHDRV_align}. Our~\model still achieves the state-of-the-art performance. The consistent performance improvement of each method also demonstrates that the pre-alignment module is reasonable.



\begin{table}[t]
	\centering
	\caption{Quantitative comparisons on RealHDRV~\cite{shu2024RealHDRV} dataset.}
	\label{tab:RealHDRV_align}
	\adjustbox{width=\linewidth}{
		\begin{tabular}{l|c|cccc}
			\toprule[1.2pt]
			\multirow{2}{*}{Type} & \multirow{2}{*}{Method} & \multicolumn{4}{c}{RealHDRV} \\
			& & TMQI$\uparrow$ & MUSIQ$\uparrow$ & PAQ2PIQ$\uparrow$ & HyperIQA$\uparrow$ \\
			\midrule \midrule
            \multirow{3}{*}{HDR Rec.} & HDR-Transformer & 0.8710 & 63.30 & 70.99 & 0.5197 \\
            & SCTNet & 0.8758 & 63.48 & 71.22 & 0.5222 \\
            & SAFNet & 0.8789 & 62.88 & 70.91 & 0.5091 \\
			\midrule
            \multirow{3}{*}{MEF} & Defusion & 0.8275 & 57.87 & 69.73 & 0.4974 \\
            & MEFLUT & 0.8505 & 62.85 & 70.93 & 0.5073 \\
            & HSDS-MEF & 0.8690 & 63.43 & 72.53 & 0.5272 \\
            \midrule
            Ours & UltraFusion & \textbf{0.8925} & \textbf{67.51} & \textbf{73.40} & \textbf{0.5833} \\
			
			\bottomrule[1.2pt]
		\end{tabular}
	}
\end{table}

\section{Discussion on MEF-SSIM}
\begin{figure}[t]
	\centering
	\includegraphics[width=\linewidth]{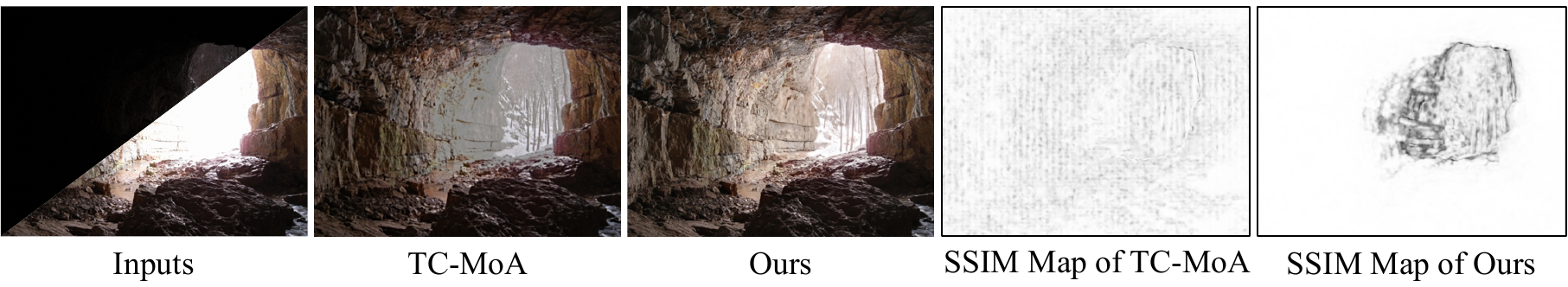}  
    \caption{Comparing MEF-SSIM map with TC-MoA~\cite{zhu2024tcmoa}.}
    \label{fig:mefssim_map}
\end{figure}

MEF-SSIM is a widely used metric to evaluate fidelity after exposure fusion. However, sometimes lower MEF-SSIM does not indicate poor fidelity. As shown in~\cref{fig:mefssim_map}, in brighter areas, ours \model achieves higher MEF-SSIM than TC-MoA, demonstrating high fidelity. In dark areas, it makes some necessary local adjustments, resulting in more natural transitions but lower MEF-SSIM. 

\section{Compare with Inapinting Methods}
To further illustrate our~\model is the first guided inpainting model that can perform artifact-free HDR imaging, we compare our method with two diffusion-based image editing methods: Anydoor~\cite{chen2024anydoor} and Stable Diffusion V2 Inpainting~\cite{stablediffusion}.

\begin{figure}[t]
	\centering
	\includegraphics[width=\linewidth]{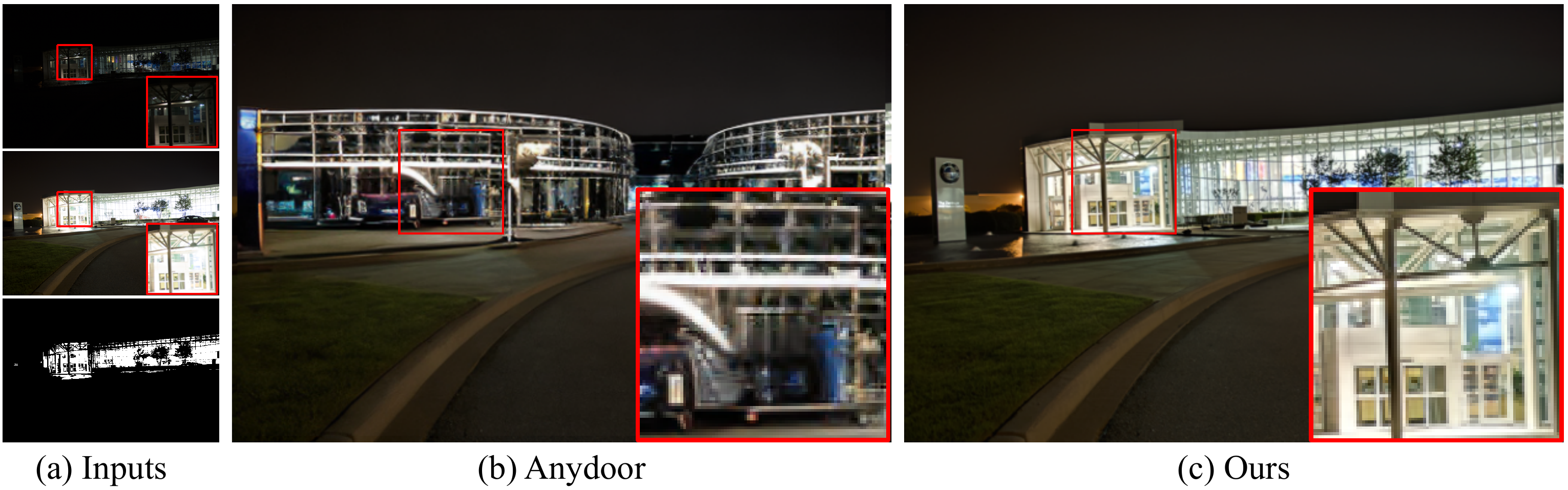}
	\caption{Compre with an image customization method Anydoor~\cite{chen2024anydoor}. Our method can preserve high-frequency details from the under-exposed image.}
	\label{fig:anydoor}
\end{figure}

\noindent\textbf{Anydoor.}
We compare our~\model with an image customization method Anydoor~\cite{chen2024anydoor}. Given a background image, a corresponding mask, and a reference image, AnyDoor can inpaint the reference into the masked region of the background image. Therefore, we utilize the over-exposed image as the background, mask out the over-exposed regions, and provide the contemporary regions from the under-exposed image as the reference.
As shown in~\cref{fig:anydoor} (b), while AnyDoor can restore the highlight regions, the restored results fail to maintain consistency with the under-exposed image. Different from Anydoor, our~\model effectively leverages the information from the under-exposed image, achieving a more reliable restoration.

\begin{figure}[!t]
	\centering
	\includegraphics[width=\linewidth]{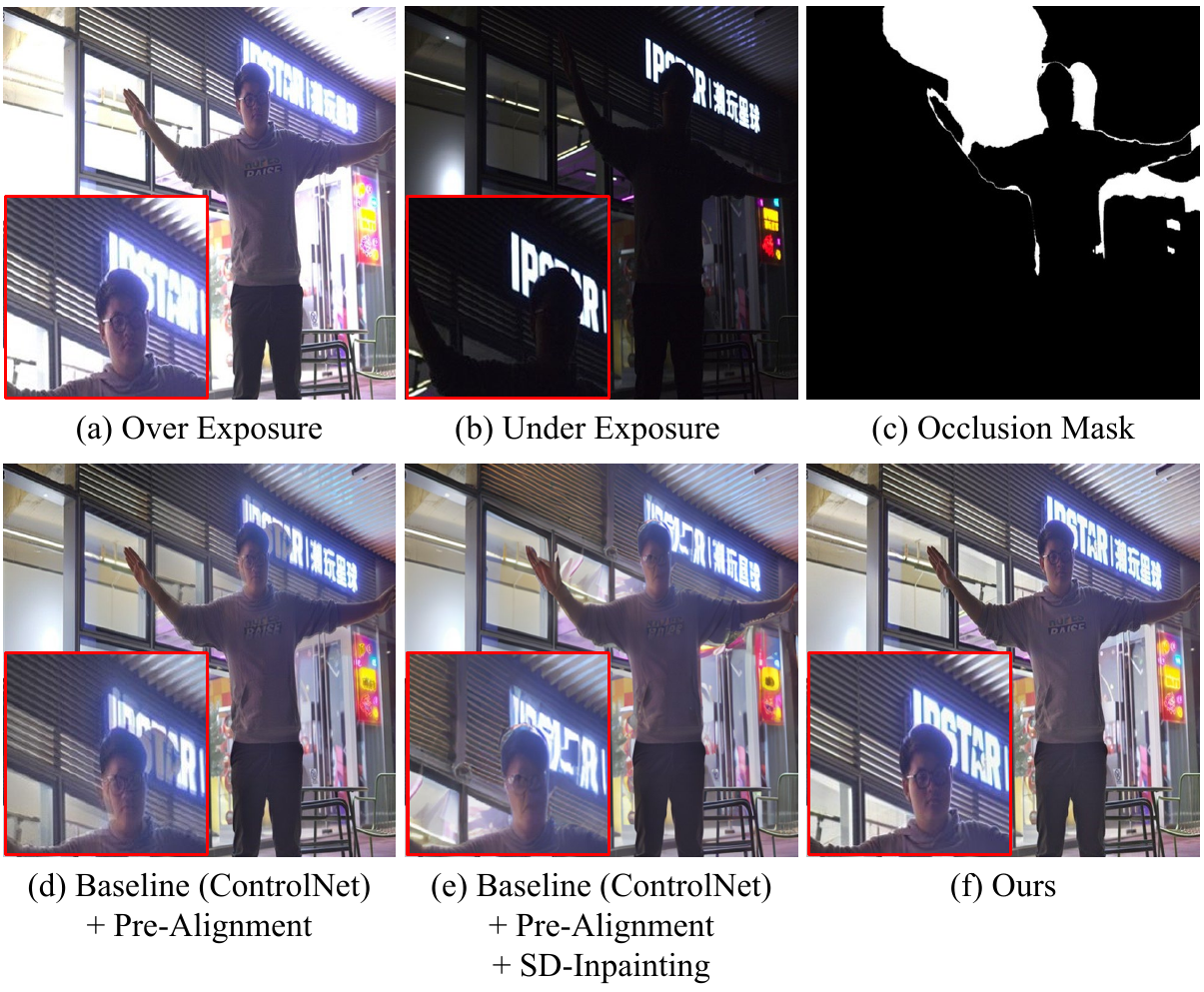}
	\caption{Visual comparisons with an inpainting method. We adopt Stable Diffusion V2 Inpainting~\cite{stablediffusion} for comparison. All the inputs are resized to $512\times512$ to meet the size requirement of the inpainting model.}
	\label{fig:inpainting}
\end{figure}

\noindent\textbf{Stable Diffusion Inpainting.}
Since Stable Diffusion V2 Inpainting~\cite{stablediffusion} lacks the ability to fuse differently exposed inputs, we first obtain an initial fused result through a pre-alignment stage and our baseline model (\ie, ControlNet~\cite{controlnet}), as shown in ~\cref{fig:inpainting} (d). Then, we use the estimated occlusion mask (\cref{fig:inpainting} (c)) as the inpainting mask for Stable Diffusion Inpainting to inpaint the occluded regions.
It can be observed from~\cref{fig:inpainting} (e) that, although the artifact effect is mitigated, due to the absence of partial under-exposed information as guidance, the result from Stable Diffusion Inpainting fails to maintain consistency with the under-exposed image. Moreover, since Stable Diffusion Inpainting is not trained on our designed synthetic data, it is not robust to align errors, leading to further distortion in well-exposed regions. Finally, without a fidelity control branch, the overall structure of the image undergoes significant deformation. In contrast, our~\model is able to generate a faithful and artifact-free output (\cref{fig:inpainting} (f)).

\section{Addtional Visual Comparisons}
We provide additional visual comparisons on three datasets (\ie, our UltraFusion benchmark, RealHDRV dataset~\cite{shu2024RealHDRV} and MEFB dataset~\cite{mefb}). Please refer to our \href{https://openimaginglab.github.io/UltraFusion/}{project page}.

For our benchmark, we present the results of our~\model and competitors on 20 scenes used for the user study. For the RealHDRV dataset, we selected 10 scenes with significant local motion. For the MEFB dataset, we randomly selected 10 scenes for visual comparison.


\end{document}